\documentclass{article}

\usepackage{PRIMEarxiv}

\usepackage[utf8]{inputenc} 
\usepackage[T1]{fontenc}    
\usepackage{hyperref}       
\usepackage{url}            
\usepackage{booktabs}       
\usepackage{amsfonts}       
\usepackage{nicefrac}       
\usepackage{microtype}      
\usepackage{lipsum}
\usepackage{color}
\usepackage{fancyhdr}       
\usepackage{graphicx}       
\graphicspath{{media/}}     

\title{Timealign: A multi-modal object detection method for time misalignment fusing in autonomous driving
}

\author{Zhihang Song,
        Lihui Peng\thanks{\textsuperscript{}Corresponding author: Lihui Peng.} ,
        Jianming Hu,
        Danya Yao,
	Yi Zhang\\
  Department of Automation \\
  Tsinghua University \\
  Beijing, China\\
  \texttt{song-zh22@mails.tsinghua.edu.cn, lihuipeng@mail.tsinghua.edu.cn} \\}

\begin{document}
\maketitle

\begin{abstract}
The multi-modal perception methods are thriving in the autonomous driving field due to their better usage of complementary data from different sensors. Such methods depend on calibration and synchronization between sensors to get accurate environmental information. There have already been studies about space-alignment robustness in autonomous driving object detection process, however, the research for time-alignment is relatively few. As in reality experiments, LiDAR point clouds are more challenging for real-time data transfer, our study used historical frames of LiDAR to better align features when the LiDAR data lags exist. We designed a Timealign module to predict and combine LiDAR features with observation to tackle such time misalignment based on SOTA GraphBEV framework.

\end{abstract}


\section{Introduction}
For autonomous driving object detection, there are many solutions based on different sensor inputs, such as camera\cite{ku2019monocular,liu2021autoshape}, LiDAR\cite{shi2019pointrcnn,lang2019pointpillars,yin2021center}, or both of them\cite{xu2018pointfusion,vora2020pointpainting,wang2021pointaugmenting,liu2023bevfusion,huang2020epnet,liu2022epnet++,li2022deepfusion}. In recent years, multi-modal fusion has emerged as a leading approach for 3D detection tasks. By leveraging complementary data from diverse sensors, such as images and point clouds, this method significantly outperforms single-sensor solutions. However, while it offers substantial advantages, multi-modal fusion introduces a set of unique challenges. A critical aspect lies in the synchronization and calibration of sensors\cite{wang2023multi,luo2023calib,xiao2023calibformer}. Without space and time alignment, effective fusion of information would be impossible. 

Recently, researchers also noticed that data misalignment may cause severe performance degradation\cite{dong2023benchmarking,graphalign} for object detection algorithms. Space misalignment and time misalignment have attracted attention because of their lack of being taken into consideration in previous algorithms design\cite{yu2023benchmarking}. As to the space alignment, a breakthrough is to use the bird's eye view (BEV) to fuse features\cite{liu2023bevfusion},\cite{cai2023bevfusion4d}, which also attracted considerable attention from researchers. Some previous work such as \cite{graphalign,song2024graphbev} has already been carried out on space alignment under bias.

As to the time alignment, previous work such as \cite{zhao20193d,cai2023bevfusion4d,huang2022bevdet4d} also noticed and utilized the sequential data for temporal fusion to further improve the detection performance, but did not aim to solve feature time misalignment caused by data synchronization. However, according to experience in real-world autonomous driving, component failures, and data inconsistencies are inevitable challenges\cite{yu2023benchmarking,wang2023multi}. Because during the data streaming process, sensor data is tagged with timestamps before being transmitted to the deep learning model through system sockets, there may be several reasons for time misalignment\cite{yu2023benchmarking,dong2023benchmarking}. Firstly, time delays may arise in the data transfer process because of sensor connection failure or temporary insufficient cable bandwidth\cite{liu2022communication}. For instance, under conditions of network congestion or hardware bottlenecks, the data frame of one modality could experience delays for seconds in severe conditions. Besides, as the LiDAR scans its surroundings in a period but the camera takes images instantaneously, the synchronization between sensor settings may cause asynchronization too\cite{dong2023benchmarking}. 

Such asynchronization poses serious risks to the accuracy and reliability of autonomous driving systems. Misaligned data frames could lead to incorrect feature associations, degraded object detection performance, or even critical errors in path planning and decision-making. To address these challenges, advanced techniques such as timestamp interpolation, sensor clock synchronization protocols, or real-time fallback strategies could be implemented\cite{liu2020computing,zhang2020sensing}. These measures ensure data consistency and system resilience in normal conditions, safeguarding the performance of autonomous driving models in complex and dynamic environments. However, timestamp accuracy is a challenging problem if different sensors with different granularities, and handling such data sources under normal and abnormal communication conditions is still an open question\cite{liu2020computing}. Thus, the perception algorithms should take these problems into consideration.

Our work uses the outstanding method GraphBEV\cite{song2024graphbev} as our base framework because it is a novel work considering space misalignment and making specific structure designs for such alignment. We are inspired by its ideas and developed our TimeAlign method aiming to improve the performance under both space and time alignment problems. Because there exists mature image compression and transmission methods\cite{jain1981image} as well as the frequency of taking instantaneous images is always higher than LiDAR scanning, and the LiDAR-stuck influences the performance more severely than the camera\cite{yu2023benchmarking}, we focus on the problem of LiDAR data lags in the perception process and take image time as the reference. GraphBEV\cite{song2024graphbev} has great robustness in space misalignment, but in our experiment of LiDAR-stuck, its performance drops significantly, indicating that feature misalignment caused by time lagging is different and can not be corrected by the same module of space alignment. Thus, we introduced a separate time alignment part in the BEV feature fusion process. Our experiments added random frame lags in LiDAR observation data. By using the historical frames of LiDAR point clouds, we utilized the Swin-LSTM structure\cite{tang2023swinlstm} to predict the BEV LiDAR feature of the current time. Then, we combined the predicted feature and the observation feature under the guidance of the image BEV feature to get more credible current LiDAR information, inspired by the idea of the Kalman filter. we also introduced corresponding LiDAR feature prediction loss into the whole training process to helps the method to converge.

The main contributions of this work include: (a) Experiments in our study support the view that data lags in LiDAR have a great impact on 3D detection tasks. (b) We proved that temporal misalignment is not consistent with space one and previous effective frameworks for space misalignment can not solve it. (c) We improved a TimeAlign model based on GraphBEV for better performance under LiDAR lagging.

This paper is organized in structures below: Section \ref{Relatedworks} introduces relative work of 3D object detection of BEV methods. In Sections \ref{Methodology}, we presented our time alignment detection framework from data preparation, structure design, and loss design in detail, respectively. In Section \ref{Experimentresults}, we conducted experiments on the baseline GraphBEV and our TimeAlign model, which proves that LiDAR lagging can't be treated and solved as space misalignment and shows our improved results. Finally, we discuss some perspectives for future research and provide a summary in Section \ref{Conclusion}.

\section{Related Works} \label{Relatedworks}
\subsection{LiDAR-Camera 3D Object Detection of BEV}

Numerous studies have been conducted in the field of 3D object detection, but here, we focus specifically on approaches utilizing multi-modal inputs from LiDAR and cameras. LiDAR point clouds excel in providing precise depth and spatial location information, offering a detailed representation of the 3D environment\cite{wang2021pointaugmenting,vora2020pointpainting}. In contrast, cameras capture rich texture and color details\cite{yu2023benchmarking}, which are indispensable for understanding visual context and object appearance. By combining these complementary data sources, it becomes possible to significantly enhance detection accuracy and robustness\cite{li2022deepfusion,xu2021fusionpainting,goodnough1999transfusion,graphalign}. As our method is based on a feature fusion method in BEV space, here we mainly introduce relative methods of such kind. 

Bird's eye view is used in 3D object detection methods nowadays and soon became one of the most promising frameworks for fusion perception. The top-down perspective of the scene provides a clear spatial layout of the environment\cite{liu2023bevfusion}, which is particularly beneficial for understanding the relationships between objects, accurately estimating distances, and planning paths in crowded or complex scenarios. Especially for autonomous driving scenes, where most targets are near the ground plane, such method simplifies the 3D information processing by projecting sensor data, such as LiDAR point clouds or camera features, into a unified 2D plane. The characteristics of BEV representation are intuitive and easy to understand for both human beings and algorithms\cite{wang2023multi}.

There are lots of high-performance previous work based on BEV, such as BEVFusion\cite{liu2023bevfusion,liang2022bevfusion}, ObjectFusion\cite{cai2023objectfusion}, MetaBEV\cite{ge2023metabev}, GraphBEV\cite{song2024graphbev} and so on\cite{wang2024unibev,song2024contrastalign,wang2024mv2dfusion,cai2023bevfusion4d}. The BEVFusion reached high performance on nuScenes Dataset and attracted attention to BEV-based detection framework but ignored the feature misalignment evaluation\cite{liu2023bevfusion,yu2023benchmarking,song2024graphbev}. The \cite{liang2022bevfusion,wang2024unibev} noticed the sensor fusion problem might occur in reality and creatively built a robust framework for sensor-missing situations. The \cite{song2024graphbev,song2024contrastalign} focused on the space feature alignment problem in fusion and used a novel alignment structure to improve the robustness of the model\cite{song2024graphbev,graphalign}.

\subsection{Temporal related 3D Object Detection}

The temporal information is an important source for environment perception and trend prediction. Using the historical frames from LiDAR is also an effective way to promote the performance of detection accuracy. For example, using historical sweeps is a normal trick in many detection methods\cite{liu2023bevfusion,song2024graphbev} to boost 3D object detection. Besides, works such as \cite{piergiovanni20214d,huang2020lstm} utilized temporal sensor frames to extract motion cues to enhance the feature and made detection more successful on Waymo Open Dataset\cite{sun2020waymo}. BEVFormer\cite{li2022bevformer} and LIFT\cite{zeng2022lift} managed to build the spatial and temporal fusion framework through the self-attention mechanism. BEVDet4D\cite{huang2022bevdet4d}, PETRv2\cite{Liu_2023_ICCV}, BEVFusion4D\cite{cai2023bevfusion4d}, extended the BEV detection framework into time domain by fusing 4D features and outperformed the responding 3D version on nuScenes Dataset\cite{caesar2020nuScenes}. The above research aimed to promote the detection performance by aggregating more temporal features. Thus, we were inspired by such ideas and designed our model for LiDAR stuck robustness using temporal information.

\section{Methodology} \label{Methodology}

This section will introduce our modification and algorithm design for time alignment based on the GraphBEV framework. We utilized the historical frames of LiDAR in the feature processing part to enhance the robustness of model under data lag. The main design of TimeALign contains two parts: LiDAR prediction and combination of prediction and observation. 

\begin{figure*}[htbp]
\centerline{\includegraphics[width=1\linewidth]{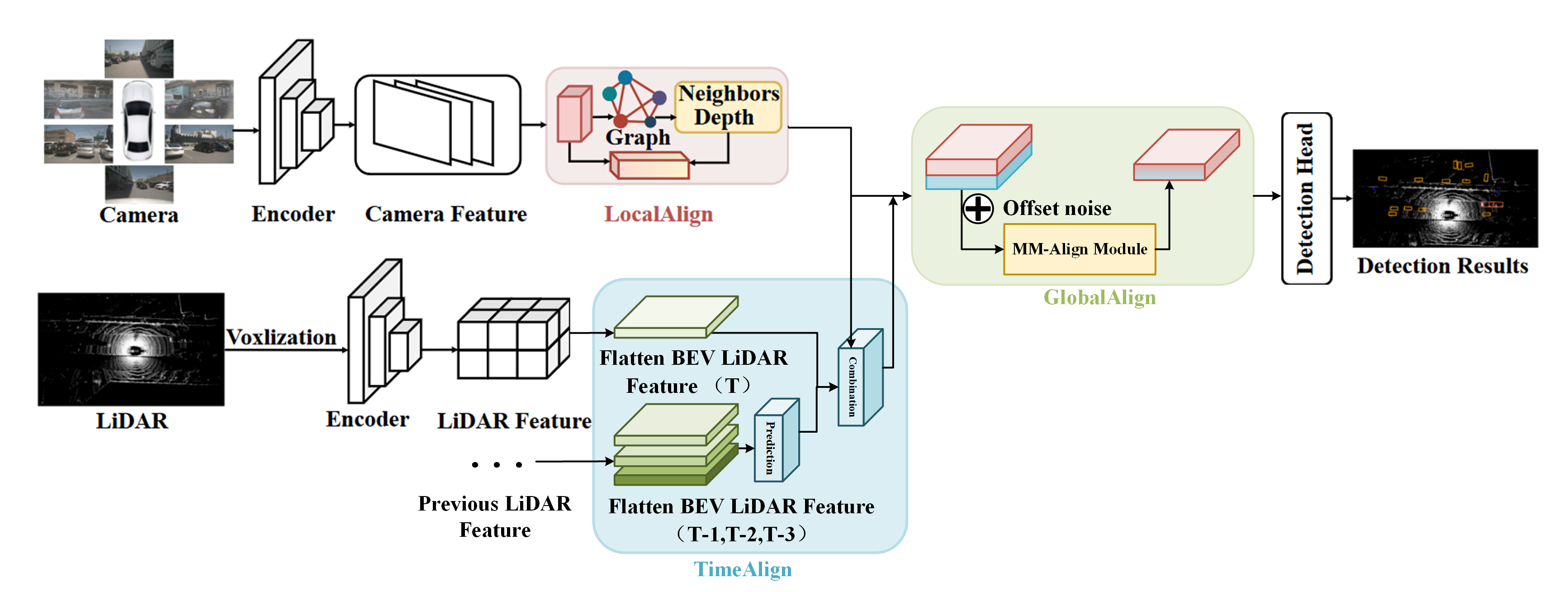}}
\caption{The whole structure of TimeAlign model.}
\label{fig: timebev}
\end{figure*}

The whole structure of our method is shown in Figure\ref{fig: timebev}. The framework is based on GraphBEV\cite{song2024graphbev}, and we modified it and added a TimeAlign module to better process temporal information. We take the six camera images at current time T and LiDAR point clouds from T-3 to T as our inputs. The previous LiDAR points and current LiDAR points will go through the voxelization and encoder modules to generate LiDAR BEV features separately. The predicted current feature will be produced with information and trend in previous frames and then be used to combine with the 'real' observed one to generate the final LiDAR BEV feature. This LiDAR final BEV feature will be futher fused with image BEV feature in GlobalAlign\cite{song2024graphbev} module and sent into detection head to get the final results.

\subsection{Data Preparation} \label{dataprepare}

We chose the nuScenes Dataset\cite{caesar2020nuScenes} as our experiment dataset for its rich information in consistent data series. We pre-processed the dataset by using tokens in nuScenes Dataset to extract three consistent historical point clouds as well as their transform matrices information and organized them for each input. For those whose timestamp is at the beginning with previous frames less than three, we use the current frame to complete. In the data input stage, the previous LiDAR frames will be projected to the corresponding ego coordinates with the transformation information organized before, and then be processed and sent into the next model with the current LiDAR frames. In this way, we got the temporal series of LiDAR information for the latter feature extraction process.

In order to imitate the LiDAR-stuck misalignment in reality, we added random LiDAR-stuck cases in our experiment. In the whole training process, we use a random variable $\alpha$ to randomly replace the current input of LiDAR point cloud with a previous frame to imitate the data lag caused by transfer congestion or sensor failure, so that the model can learn how to process information both in normal and corrupted conditions.

\subsection{LiDAR BEV Feature Prediction} 

The TimeAlign module mainly completes two works. One is to get predicted LiDAR features from historical information. The other is to utilize both the current observed feature and the predicted feature to learn the final current LiDAR feature which is corresponding to the timestamp of the camera ones.

For the prediction part, we referred to the novel prediction method Swin-LSTM\cite{tang2023swinlstm}. In this part, the series of previous LiDAR BEV feature is fed into a recurrent LSTM structure based on a multi-head SwinTransformer. The structure is shown below in Figure\ref{fig: prediction}. Such method combines the strengths of SwinTransformer and Long Short-Term Memory networks to model spatial and temporal dependencies effectively. The recurrent structure contains a patch embedding layer, Swin-LSTM cells, and a patch inflation layer. The patch embedding layer splits input BEV features into patches and embedding them into a high-dimensional feature space. The Swin-LSTM cells utilize the idea of SwinTransformer, which is based on self-attention within local windows to capture fine-grained spatial features. In the SwinLSTM cells, we used the depth multi-head STB modules\cite{tang2023swinlstm} to generate the prediction features and states. The patch inflation layer reconstructs the prediction feature from hidden states and outputs from cells. 

To predict LiDAR features from historical series, We first input the earliest previous frame to generate the next time step feature and hidden state, then we used the output of the last step as the input to the prediction model to obtain the next step recurrently, until reaching the current time step. In our experiment, we chose the input series length of 3, so that we could develop the predicted features from T-2 to T. These predicted features will be used to calculate MSE (Mean Square Error) loss with each corresponding time step label feature to train the prediction model.

\begin{figure*}[htbp]
\centerline{\includegraphics[width=0.5\linewidth]{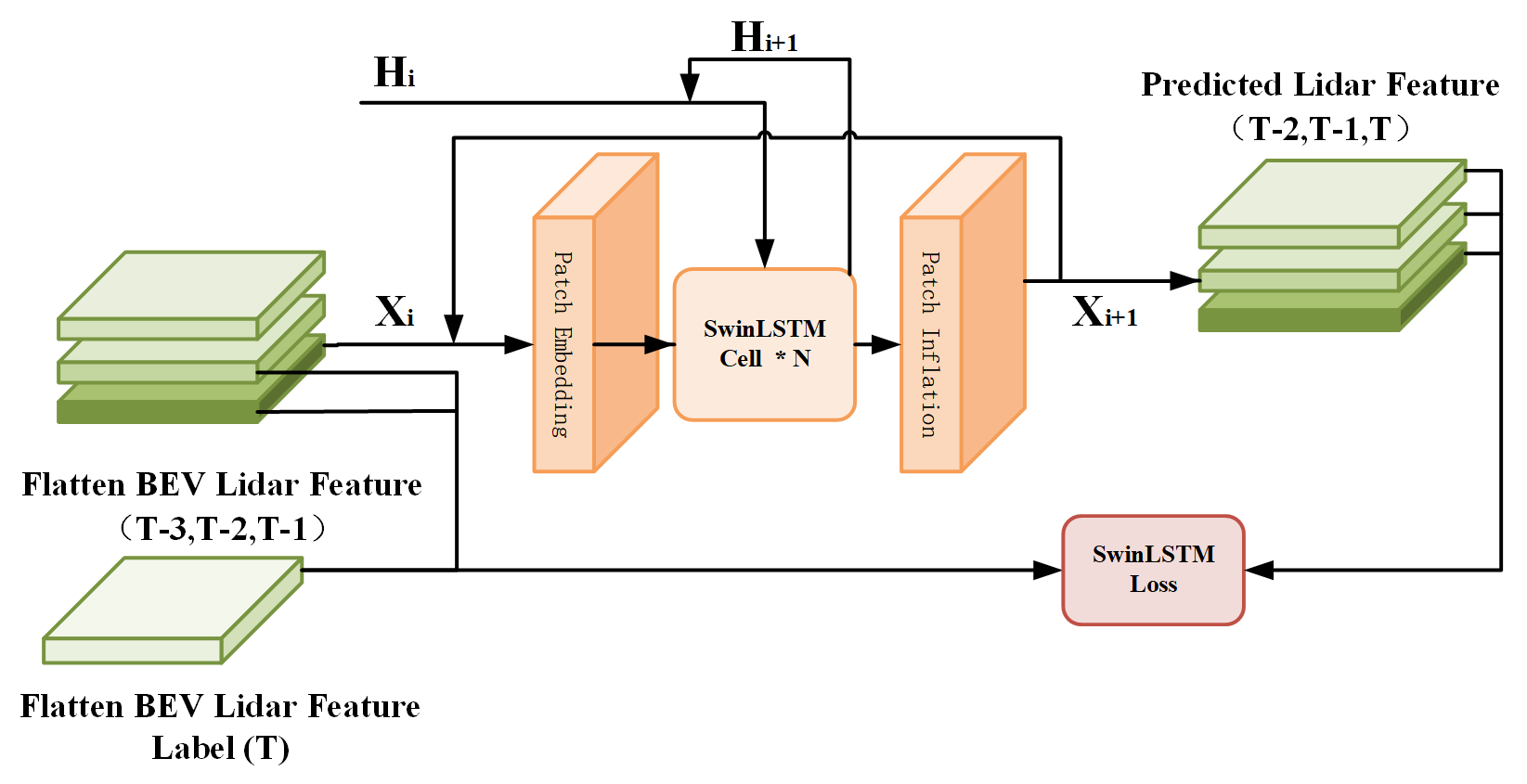}}
\caption{Structure of Swin-LSTM LiDAR feature prediction module in TimeAlign.}
\label{fig: prediction}
\end{figure*}

\subsection{Prediction and Observation Combination} 

After we got the predicted LiDAR BEV features from historical frames, the next problem we would like to solve is how to combine the predicted features with the observed ones. As our experiment settings imitated cases with randomly delayed observation, we need to make the model learn whether the observation is temporal trustworthy, which means if it is the LiDAR feature corresponding to the moment when the cameras capture. Thus, we design a dual-transformer module to combine the prediction and the observation under the guidance of camera BEV features. The structure of the combination part is shown below in Figure\ref{fig: combination}.

\begin{figure*}[htbp]
\centerline{\includegraphics[width=0.5\linewidth]{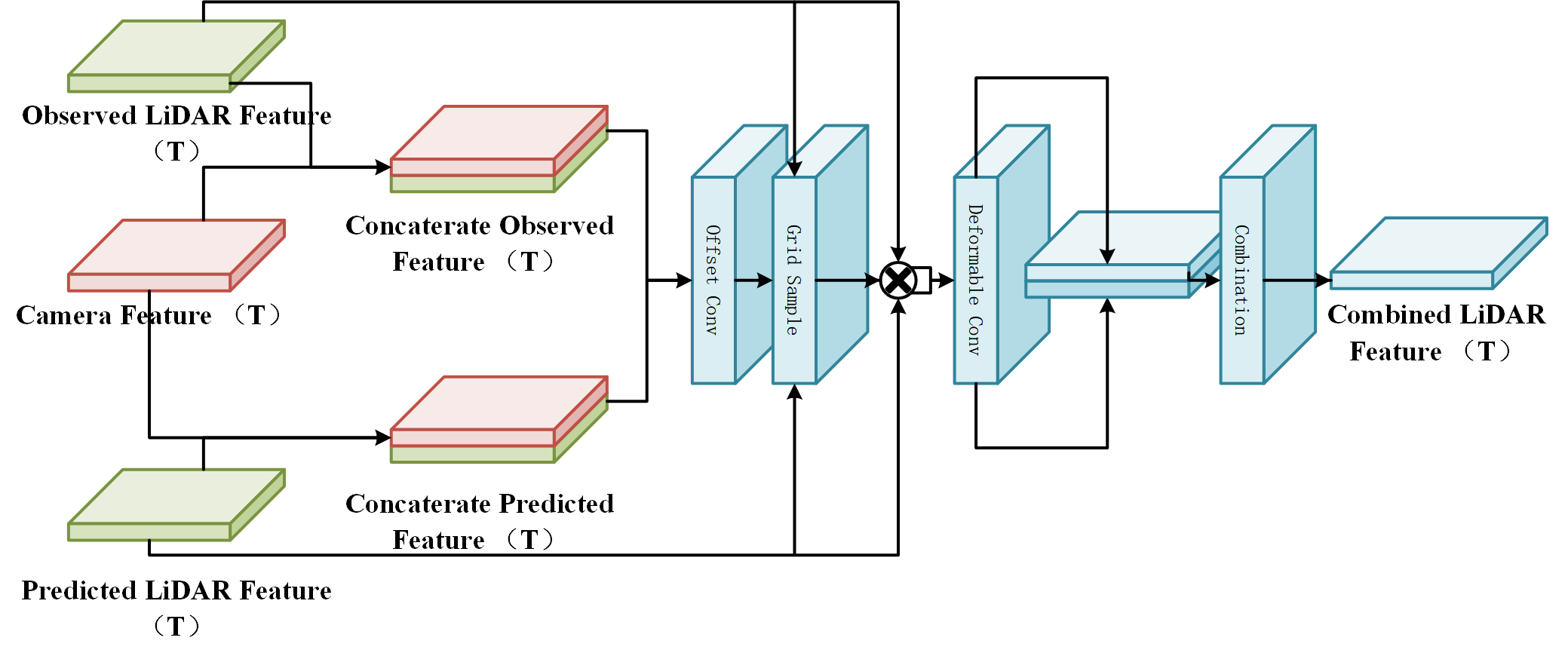}}
\caption{Structure of combination layers of predicted LiDAR feature and observed ones in TimeAlign.}
\label{fig: combination}
\end{figure*}

To determine whether the prediction or the observation of LiDAR features is temporally aligned needs the reference of the current time, so the camera feature is used here to evaluate the alignment and guide the calculation of offsets. The predicted features $F_p \mathbb{\in} \mathbb{R}^{B*256*180*180}$ and observed features $F_o \mathbb{\in} \mathbb{R}^{B*256*180*180}$ of LiDAR are both concatenated with camera BEV features $F_c \mathbb{\in} \mathbb{R}^{B*80*180*180}$ separately. Then, we used the concatenated features $F_{combine}^{pre} \mathbb{\in} \mathbb{R}^{B*336*180*180}, F_{combine}^{ob} \mathbb{\in} \mathbb{R}^{B*336*180*180}$ to calculate the learnable offset parameters between the LiDAR and camera features in the offset convolution module. These parameters represent the misalignment between sensors and guide the grid-sample layer to obtain deform weights for LiDAR features. Then we use the deform weights to multiple previous features as the input of the deformable convolution module. This module dynamically generates a new aligned feature to adjust the bias caused by the movements of objects in lagging intervals. Such a process is conducted on both predicted and observed features to obtain the re-aligned results separately and then we fuse the observed and predicted results in combination layers to get the final LiDAR BEV features $F_f \mathbb{\in} \mathbb{R}^{B*256*180*180}$.

\section{Experiment Results} \label{Experimentresults}
In our experiment, we chose the nuScenes Dataset as our training and testing dataset. The nuScenes dataset\cite{caesar2020nuScenes} is a comprehensive large-scale benchmark for 3D detection tasks, comprising 700 scenes for training, 150 scenes for validation, and 150 scenes for testing. The dataset was captured using six multi-view cameras and a 32-channel LiDAR sensor, providing extensive coverage and diverse perspectives. It features 360-degree annotations across 10 object classes, making it a valuable resource for evaluating detection algorithms. The primary evaluation metric used to assess detection performance is the mean Average Precision (mAP). We used the full train set of nuScenes Dataset with 28130 samples under 700 scenes and the mini test set for quick testing. Our method based on the code of GraphBEV\cite{song2024graphbev}, BEVFusion\cite{liu2023bevfusion}, OpenPCDet\cite{od2020openpcdet}, and OpenSTL\cite{tan2023openstl}. Our method inherited the structure of BEVFusion, using SwinTransformer\cite{liu2021swin} to extract image features and SECOND\cite{yan2018second} for the LiDAR modal. LSS\cite{philion2020lift} is also used for image BEV feature projection. OpenSTL\cite{tan2023openstl} is an open source benchmark for spacial and temporal predictive methods and here we developed our time alignment module based on it.

Firstly, we test the performance of GraphBEV under 1 LiDAR frame lagging to explore whether the spacial alignment method in GraphBEV can either handle the temporal alignment. We trained the GraphBEV in the Nuscense Dataset with settings of K=8 and sweeps=1. The test results of 4 detection object types are shown below in Table\ref{tab1}. From the table, we can see that the performance of all AP in different types of objects drops apparently, which means that the spacial alignment method failed in the temporal misalignment. The reason might be that the spacial misalignment caused by calibration bias is an affine transformation bias, however, it is structurally different from the temporal cases. The misalignment caused by LiDAR stuck is mainly generated from the independent movements of the ego car and the surrounding objects, and thus is more complicated and can not be represented by a single affine transformation. So to design a specific method for such misalignment is of value.

\begin{table}[htbp]
\begin{center}
\caption{Test of GraphBEV performance with and without temporal misalignment (1 LiDAR frame lagging).}
\label{tab1}

\begin{tabular}{l*{5}{c}}
\hline

\textbf{\textit{Situation}} & 
\textbf{\textit{AP-Car}}
&\textbf{\textit{AP-Truck}}& \textbf{\textit{AP-Bus}}& \textbf{\textit{AP-Pedestrian}}&\\

\hline
GraphBEV LiDAR(T)&0.823&0.595&0.980&0.858 \\
GraphBEV LiDAR Lagging(T-1) &0.595&0.477&0.306&0.613 \\

\hline
\end{tabular}
\end{center}
\end{table}

In our experiment, we developed the Timealign method based on GraphBEV and kept the same basic settings of K=8 and sweeps =1. In our training, We also used the pre-trained model checkpoint of GrapBEV and loaded the same layers into our model to make the training converge more quickly and easily. And we also set our training in two steps, with the loss of a combination of prediction loss and GraphBEV detection loss accounting for different proportions. Our experiment set the weight of prediction loss is 10 and 0.001. After 25 epochs of training, the Timealign model converged and the test results on mini test set are shown below in Table\ref{tab2}. Here we also listed the results from GraphBEV in table to compare.

\begin{table}[htbp]
\begin{center}
\caption{Comparison of Timealign performance with and without temporal misalignment (1 LiDAR frame lagging).}
\label{tab2}

\begin{tabular}{l*{5}{c}}
\hline

\textbf{\textit{Situation}} & 
\textbf{\textit{AP-Car}}
&\textbf{\textit{AP-Truck}}& \textbf{\textit{AP-Bus}}& \textbf{\textit{AP-Pedestrian}}&\\

\hline
GraphBEV LiDAR(T)&0.823&0.595&0.980&0.858 \\
GraphBEV LiDAR Lagging(T-1) &0.595&0.477&0.306&0.613 \\
Timealign LiDAR(T)&0.789&0.597&0.927&0.840 \\
Timealign LiDAR Lagging(T-1) &0.656&0.513&0.769&0.759 \\

\hline
\end{tabular}
\end{center}
\end{table}

From the results, we can see that under the condition with LiDAR stuck, though the AP of 4 types of objects also degrades with corrupted data, the performance of Timealign has an improvement compared to GraphBEV. The feature prediction and combination modules utilized the historical information and made the detection process more robust under such time misalignment. 

We also noticed that the performance of Timealign appears to have a slight drop than GraphBEV under correct data. We acknowledged that is the limit of our method. We analyzed that it may be because the prediction of the current timestep is not accurate enough. In the combination part, such prediction errors are also introduced into the final LiDAR feature but are not filtered clearly by the true current observation inputs. Besides, our model is based on the three historical LiDAR inputs limited by the hardware memory size and didn't conduct experiment on more dense and longer frame series, which may also lead to the accuracy limitation. Thus, this will be the problem we mainly focus on in our future research, and we hope that our developed ideal method can better utilize the historical information to improve performance under both conditions. We will learn more from the BEVFusion4D\cite{cai2023bevfusion4d}, BEVDet4D\cite{huang2022bevdet4d} and more relative works to polish our design.

\section{Conclusion} \label{Conclusion}
In this paper, we introduced a multi-model 3D object detection method for temporal misalignment issues.  
In this study, we introduce Timealign, a multi-model 3D object detection method designed to tackle the temporal misalignment challenges inherent in BEV-based methods. To address temporal misalignment caused by data lagging in LiDAR, we propose the Timelign module, which utilizes the historical LiDAR features through Swin-LSTM structure to predict the current states. Additionally, to resolve the alignment during the fusion of predicted and observed LiDAR BEV features corresponding to the camera features at current time, we develop the dual-trans method to match and refine the combination. This module simulates the LiDAR stuck by random frame drop and subsequently aligns multi-modal features using learnable offsets, ensuring temporal integration of information from multiple sensors.


\bibliographystyle{unsrt}  
\bibliography{references}

\end{document}